\documentclass[letterpaper]{article} 
\usepackage{aaai2026}  
\usepackage{times}  
\usepackage{helvet}  
\usepackage{courier}  
\usepackage[hyphens]{url}  
\usepackage{graphicx} 
\usepackage{natbib}  
\usepackage{caption} 
\usepackage{algorithm}
\usepackage{algorithmic}

\usepackage{amsmath}
\usepackage{amssymb}
\usepackage{booktabs}
\usepackage[dvipsnames]{xcolor}

\definecolor{White}{rgb}{1.,0.,1.}
\definecolor{first}{rgb}{.8,.0,.0}
\definecolor{second}{rgb}{.0,.6,.0}
\definecolor{third}{rgb}{.0,.0,.8}

\definecolor{ceiling}{RGB}{214,  38, 40}
\definecolor{floor}{RGB}{43, 160, 4}
\definecolor{wall}{RGB}{158, 216, 229}
\definecolor{window}{RGB}{114, 158, 206}
\definecolor{chair}{RGB}{204, 204, 91}
\definecolor{bed}{RGB}{255, 186, 119}
\definecolor{sofa}{RGB}{147, 102, 188}
\definecolor{table}{RGB}{30, 119, 181}
\definecolor{tvs}{RGB}{160, 188, 33}
\definecolor{furniture}{RGB}{255, 127, 12}
\definecolor{objects}{RGB}{196, 175, 214}

\definecolor{car}{rgb}{0.39215686, 0.58823529, 0.96078431}
\definecolor{bicycle}{rgb}{0.39215686, 0.90196078, 0.96078431}
\definecolor{motorcycle}{rgb}{0.11764706, 0.23529412, 0.58823529}
\definecolor{truck}{rgb}{0.31372549, 0.11764706, 0.70588235}
\definecolor{othervehicle}{rgb}{0.39215686, 0.31372549, 0.98039216}
\definecolor{person}{rgb}{1.        , 0.11764706, 0.11764706}
\definecolor{bicyclist}{rgb}{1.        , 0.15686275, 0.78431373}
\definecolor{motorcyclist}{rgb}{0.58823529, 0.11764706, 0.35294118}
\definecolor{road}{rgb}{1.        , 0.        , 1.        }
\definecolor{parking}{rgb}{1.        , 0.58823529, 1.        }
\definecolor{sidewalk}{rgb}{0.29411765, 0.        , 0.29411765}
\definecolor{otherground}{rgb}{0.68627451, 0.        , 0.29411765}
\definecolor{building}{rgb}{1.        , 0.78431373, 0.        }
\definecolor{fence}{rgb}{1.        , 0.47058824, 0.19607843}
\definecolor{vegetation}{rgb}{0.        , 0.68627451, 0.        }
\definecolor{trunk}{rgb}{0.52941176, 0.23529412, 0.        }
\definecolor{terrain}{rgb}{0.58823529, 0.94117647, 0.31372549}
\definecolor{pole}{rgb}{1.        , 0.94117647, 0.58823529}
\definecolor{trafficsign}{rgb}{1.        , 0.        , 0.        }
\definecolor{otherstructure}{rgb}{0.98039215, 0.58823529, 0.}
\definecolor{otherobject}{rgb}{0.19607843, 1.        , 1.        }

\makeatletter
\newcommand{\car@semkitfreq}{3.92}
\newcommand{\bicycle@semkitfreq}{0.03}
\newcommand{\motorcycle@semkitfreq}{0.03}
\newcommand{\truck@semkitfreq}{0.16}
\newcommand{\othervehicle@semkitfreq}{0.20}
\newcommand{\person@semkitfreq}{0.07}
\newcommand{\bicyclist@semkitfreq}{0.07}
\newcommand{\motorcyclist@semkitfreq}{0.05}
\newcommand{\road@semkitfreq}{15.30}
\newcommand{\parking@semkitfreq}{1.12}
\newcommand{\sidewalk@semkitfreq}{11.13}
\newcommand{\otherground@semkitfreq}{0.56}
\newcommand{\building@semkitfreq}{14.1}
\newcommand{\fence@semkitfreq}{3.90}
\newcommand{\vegetation@semkitfreq}{39.3}
\newcommand{\trunk@semkitfreq}{0.51}
\newcommand{\terrain@semkitfreq}{9.17}
\newcommand{\pole@semkitfreq}{0.29}
\newcommand{\trafficsign@semkitfreq}{0.08}
\newcommand{\semkitfreq}[1]{{\csname #1@semkitfreq\endcsname}}

\newcommand{\car@sscbkitfreq}{2.85}
\newcommand{\bicycle@sscbkitfreq}{0.01}
\newcommand{\motorcycle@sscbkitfreq}{0.01}
\newcommand{\truck@sscbkitfreq}{0.16}
\newcommand{\othervehicle@sscbkitfreq}{5.75}
\newcommand{\person@sscbkitfreq}{0.02}
\newcommand{\road@sscbkitfreq}{14.98}
\newcommand{\parking@sscbkitfreq}{2.31}
\newcommand{\sidewalk@sscbkitfreq}{6.43}
\newcommand{\otherground@sscbkitfreq}{2.05}
\newcommand{\building@sscbkitfreq}{15.67}
\newcommand{\fence@sscbkitfreq}{0.96}
\newcommand{\vegetation@sscbkitfreq}{41.99}
\newcommand{\terrain@sscbkitfreq}{7.10}
\newcommand{\pole@sscbkitfreq}{0.22}
\newcommand{\trafficsign@sscbkitfreq}{0.06}
\newcommand{\otherstructure@sscbkitfreq}{4.33}
\newcommand{\otherobject@sscbkitfreq}{0.28}
\newcommand{\sscbkitfreq}[1]{{\csname #1@sscbkitfreq\endcsname}}

\usepackage{newfloat}
\usepackage{listings}
\DeclareCaptionStyle{ruled}{labelfont=normalfont,labelsep=colon,strut=off} 
\floatstyle{ruled}
\newfloat{listing}{tb}{lst}{}
\floatname{listing}{Listing}
\title{Unleashing Semantic and Geometric Priors for 3D Scene Completion}
\author{
    Shiyuan Chen\textsuperscript{\rm 1}\thanks{Equal contribution. Work done during Shiyuan Chen's internship at D-Robotics.},
    Wei Sui\textsuperscript{\rm 2}\footnotemark[1],
    Bohao Zhang\textsuperscript{\rm 2},
    Zeyd Boukhers\textsuperscript{\rm 3},
    John See\textsuperscript{\rm 4},
    Cong Yang\textsuperscript{\rm 1}\thanks{Corresponding author.} 
}
\affiliations{
    \textsuperscript{\rm 1}School of Future Science and Engineering, Soochow University, Suzhou, China
    \\
    \textsuperscript{\rm 2}D-Robotics, Beijing, China
    \\
    \textsuperscript{\rm 3}Fraunhofer Institute for Applied Information Technology, Sankt Augustin, Germany
    \\
    \textsuperscript{\rm 4}School of Mathematical and Computing Sciences, Heriot-Watt University Malaysia, Putrajaya, Malaysia
    
    sychen66@stu.suda.edu.cn, \{wei.sui, bohao.zhang\}@d-robotics.cc, \\ zeyd.boukhers@fit.fraunhofer.de, J.See@hw.ac.uk, cong.yang@suda.edu.cn
}

\usepackage{bibentry}

\begin{document}

\maketitle

\begin{abstract}
Camera-based 3D semantic scene completion (SSC) provides dense geometric and semantic perception for autonomous driving and robotic navigation. However, existing methods rely on a coupled encoder to deliver both semantic and geometric priors, which forces the model to make a trade-off between conflicting demands and limits its overall performance. To tackle these challenges, we propose FoundationSSC, a novel framework that performs dual decoupling at both the source and pathway levels. At the source level, we introduce a foundation encoder that provides rich semantic feature priors for the semantic branch and high-fidelity stereo cost volumes for the geometric branch. At the pathway level, these priors are refined through specialised, decoupled pathways, yielding superior semantic context and depth distributions. Our dual-decoupling design produces disentangled and refined inputs, which are then utilised by a hybrid view transformation to generate complementary 3D features. Additionally, we introduce a novel Axis-Aware Fusion (AAF) module that addresses the often-overlooked challenge of fusing these features by anisotropically merging them into a unified representation. Extensive experiments demonstrate the advantages of FoundationSSC, achieving simultaneous improvements in both semantic and geometric metrics, surpassing prior bests by +0.23 mIoU and +2.03 IoU on SemanticKITTI. Additionally, we achieve state-of-the-art performance on SSCBench-KITTI-360, with 21.78 mIoU and 48.61 IoU.
\end{abstract}

\begin{links}
  \link{Code}{https://github.com/D-Robotics-AI-Lab/FoundationSSC}
\end{links}

\section{Introduction}
\label{sec:intro}
Semantic Scene Completion (SSC) infers the complete geometry and semantics of a 3D scene from partial observations. It is essential for high-level 3D perception applications, such as intelligent driving~\cite{zheng2024occworld} and robotic navigation~\cite{wu2024embodiedocc}. Early SSC methods primarily utilised LiDAR input~\cite{song2017sscnet, roldao2020lmscnet}, but they faced challenges due to high hardware costs and limited resolution. As a result, there has been a recent shift towards camera-based SSC methods, which are gaining significant attention for their lower costs and richer texture information.

\begin{figure}[t!]
  \centering
    \includegraphics[width=0.98\linewidth]{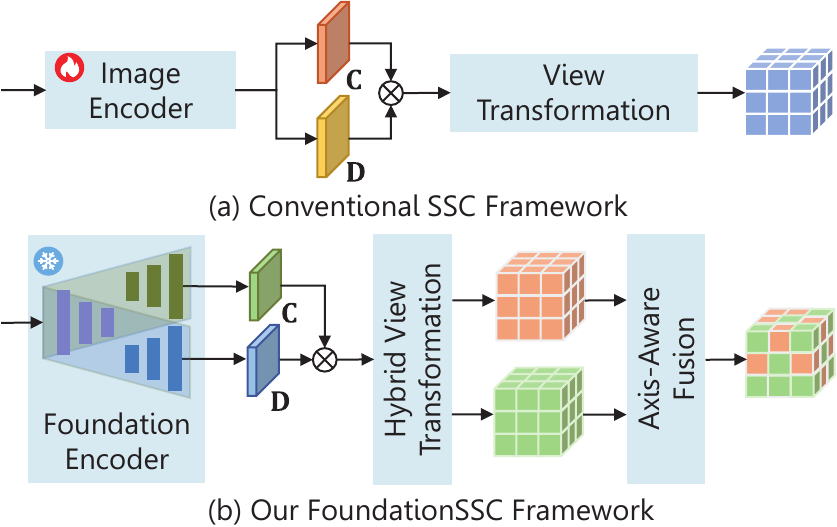}
    \caption{The conventional SSC framework (top) depends on a coupled image encoder, providing limited feature priors for both the semantic and geometric branches, resulting in an inherent trade-off. In contrast, our FoundationSSC (bottom) utilises a foundation encoder that provides robust decoupled semantic and geometric priors, effectively addressing conflicts at both the source and pathway levels.}
    \label{fig:arch_compare}
\end{figure}

Despite significant advancements, existing camera-based SSC methods are fundamentally limited by their dependence on a single, coupled source of features, as shown in Fig.~\ref{fig:arch_compare} (a). This restriction creates an inherent trade-off between the conflicting demands of semantic and geometric reasoning, ultimately hindering overall performance. Some research has attempted to address these issues by incorporating additional priors. For example, certain studies have aimed to enhance geometric understanding by utilising external stereo depth information~\cite{li2023BRGScene}. However, these approaches often function in a coarse and lossy manner, failing to fully capitalise on the rich probabilistic cues available in the cost volume. 
Conversely, other studies focus on integrating semantic priors from instance-level information~\cite{xiao2024iamssc} or Vision-Language Models (VLMs)~\cite{wang2025vlscene}. While this approach can benefit segmentation, it introduces a strong semantic bias into the coupled encoder, causing it to neglect the precise spatial features required for geometric reasoning and thus worsening the trade-off. Overall, these methods remain limited, as they merely add external inputs to a coupled framework without addressing the underlying feature conflict. Fortunately, recent advances in Vision Foundation Models (VFMs)~\cite{oquab2024dinov2} present a promising opportunity. These models exhibit remarkable zero-shot generalisation on downstream tasks, such as semantic segmentation or depth estimation. Nonetheless, we believe that simply replacing the image encoder with VFMs is not enough. The primary challenge lies in designing a novel framework that can effectively leverage these robust priors to resolve the coupling problem.

In this paper, we propose FoundationSSC, a novel SSC framework designed to unleash the semantic and geometric priors from foundation models, as shown in Fig.~\ref{fig:arch_compare} (b). Our approach operates on a core principle of dual decoupling. It begins with a unified foundation encoder that harnesses the power of VFMs to provide separate priors for the semantic and geometric branches, thereby addressing the trade-off at the feature source. These distinct priors are then processed through specialised, decoupled pathways aimed at maximising their potential. The resulting high-quality 2D representations are transformed into 3D space using a hybrid view transformation. Additionally, we introduce a novel Axis-Aware Fusion (AAF) module, which fuses the complementary 3D feature volumes in a way that respects their anisotropic characteristics, resulting in a unified 3D representation. Through these integrated designs, FoundationSSC effectively resolves the semantic-geometric coupling issues that have limited previous work, producing high-quality 2D and 3D feature representations. Consequently, our framework achieves state-of-the-art performance on SemanticKITTI~\cite{behley2019semantickitti} and SSCBench-KITTI-360~\cite{liao2022kitti360} datasets, simultaneously enhancing both geometric (Intersection over Union, IoU) and semantic (mean IoU, mIoU) metrics. In summary, our contributions are as follows:

\begin{itemize}
     \item We propose a novel framework that decouples semantics and geometric reasoning at both the source and pathway levels. This design fundamentally resolves the inherent trade-off of existing methods.
    \item We design an Axis-Aware Fusion (AAF) module that effectively fuses complementary 3D feature volumes from the hybrid view transformation, which are typically used in isolation or discarded by prior work, by respecting the anisotropic structure of driving scenes.
    \item The proposed FoundationSSC achieves state-of-the-art performance with a mIoU of 19.32 and an IoU of 48.12 on SemanticKITTI, as well as a mIoU of 21.78 and an IoU of 48.61 on SSCBench-KITTI-360, demonstrating the significant advantages of our decoupling design.
\end{itemize}
\section{Related Work}
\label{sec:related}
\subsection{Vision Foundation Models}
VFMs are trained on large-scale web datasets, enabling them to learn highly generalisable visual representations with impressive zero-shot capabilities. One of the early influential models, Contrastive Language-Image Pre-training (CLIP)~\cite{radford2021clip}, learns joint vision-language embeddings through contrastive learning. Subsequent models, like the Segment Anything Model (SAM)~\cite{kirillov2023sam, ravi2024sam2}, offer a universal segmentation framework; however, their features are not optimised for general-purpose representation learning. To address this limitation, the DINO series~\cite{caron2021dino, oquab2024dinov2} employs self-supervised learning to produce dense, high-quality features that transfer effectively to downstream tasks such as detection and segmentation. This success has paved the way for geometry-focused VFMs. The DepthAnything family~\cite{yang2024depthanything, yang2024depthanythingv2} fine-tunes DINOv2 for robust monocular depth estimation. More recently, FoundationStereo~\cite{wen2025foundationstereo} has extended this approach to stereo matching, achieving strong zero-shot generalisation across diverse domains. Inspired by these advancements, some recent SSC works have leveraged the priors of VFMs to enhance semantic features~\cite{cui2025loma, wang2025vlscene}. However, their exploration into geometry-focused foundation models remains limited.


\begin{figure*}[t!]
    \includegraphics[width=\textwidth]{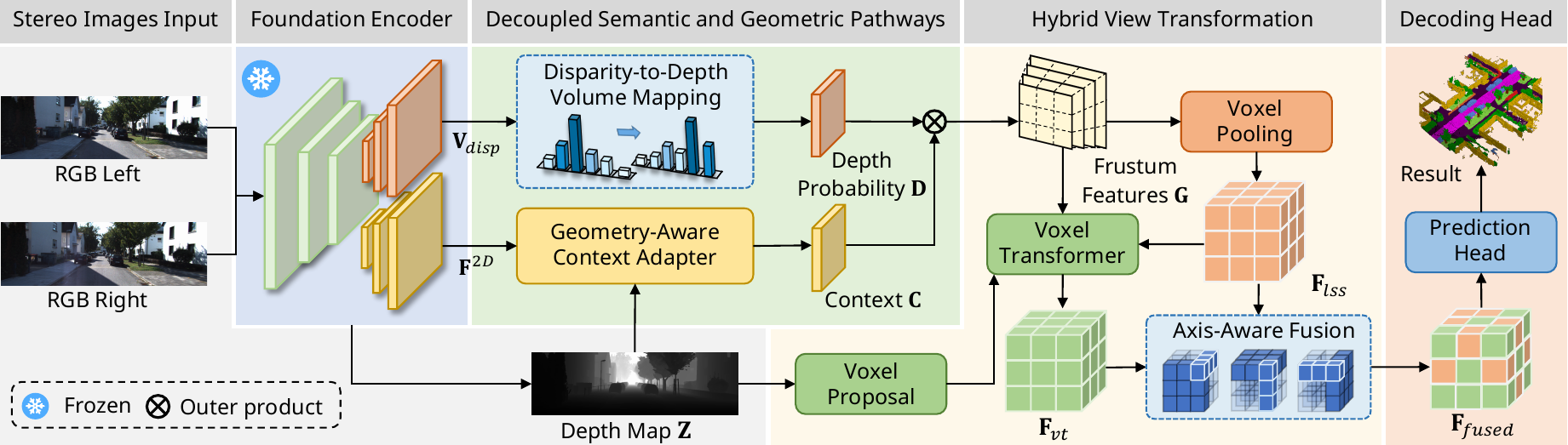}
    \caption{Overview of our proposed FoundationSSC framework. It begins with a Foundation Encoder producing decoupled priors, which are then enhanced in Decoupled Semantic Geometric Pathways. A Hybrid View Transformation subsequently lifts and fuses these priors into a unified 3D volume, which is processed by a Decoding Head to yield the final prediction.}
    \label{fig:overview}
\end{figure*}

\subsection{Camera-Based Semantic Scene Completion}
MonoScene~\cite{cao2022monoscene} is a pioneering effort that extends the SSC task to cameras by employing a Features Line of Sight Projection (FLoSP) for projecting 2D features into 3D space. TPVFormer~\cite{huang2023TPVFormer} introduces an efficient tri-perceptive view representation, created from image features. VoxFormer~\cite{li2023voxformer} employs a sparse-to-dense strategy, which propagates image features to dense voxels through deformable attention mechanism. OccFormer~\cite{zhang2023occformer} builds upon the Lift-Splat-Shoot (LSS) paradigm to facilitate feature lifting using both local and global transformer pathways. 

Subsequent works have proposed more specialised solutions, often leveraging temporal cues from multiple frames~\cite{li2024htcl, lee2025soap}, utilising more accurate geometric information obtained from stereo matching~\cite{li2023BRGScene, xue2024bissc, yu2024CGFormer}, exploiting spatial biases~\cite{wang2024h2gformer, bae2025scanssc}, and integrating richer semantic knowledge derived from instance-level features~\cite{jiang2024symphonize} or vision-language models~\cite{wang2025vlscene, cui2025loma}. However, despite the diversity of these approaches, they are often limited by a tightly coupled architecture, which results in a trade-off between semantic and geometric features. 

Our proposed FoundationSSC disentangles semantic and geometric features at both the source and processing pathway levels, effectively resolving the trade-off that has constrained earlier work.
\section{Methodology}
\label{sec:method}
As illustrated in Fig.~\ref{fig:overview}, FoundationSSC starts with a Foundation Encoder that achieves source-level decoupling, producing distinct semantic features and geometric cost volume features. These priors are then enhanced through Decoupled Semantic and Geometric Pathways: the semantic features are integrated with 3D structural awareness, while the cost volumes are transformed into rich probabilistic depth distributions. Following this, a Hybrid View Transformation elevates these refined 2D representations into two complementary 3D feature volumes. Our proposed Axis-Aware Fusion module then combines these volumes anisotropically into a unified 3D representation, which is ultimately passed to a decoding head to generate the final prediction.
\subsection{Foundation Encoder}
To address the balance between semantic and geometric information at the feature source, we utilise the pre-trained, frozen FoundationStereo model~\cite{wen2025foundationstereo} as an integrated foundation encoder. This model effectively provides reliable, separate semantic and geometric priors. This choice is particularly advantageous because FoundationStereo inherits valuable semantic features from its DINOv2 lineage~\cite{oquab2024dinov2, yang2024depthanythingv2} and also offers strong geometric reasoning based on stereo input. Consequently, we have a robust encoder that generates intrinsically separated semantic and geometric features, displaying impressive zero-shot generalisation without the need for task-specific fine-tuning. From this frozen encoder, we extract three essential outputs for our decoupled pathways:
\begin{itemize}
    \item \textbf{Monocular image features $\mathbf{F}^{2D}$:} Multi-scale features from the frozen DepthAnythingV2 backbone are fused via a lightweight Feature Pyramid Network (FPN) to produce a single-scale, robust 2D feature map, which serves as the primary input for the semantic context branch.
    \item \textbf{Cost volume features $\mathbf{V}_{disp}$:} The disparity probability distribution is generated by FoundationStereo's attentive hybrid cost filtering module. The cost volume features provide rich geometric priors for our depth branch, avoiding the information loss of collapsed depth maps.
    \item \textbf{Dense depth map $\mathbf{Z}$:} The final disparity map predicted by FoundationStereo, which is then converted to a metric depth map using camera intrinsics. This is primarily used for generating voxel proposals in the view transformation and geometric prior for the semantic context branch.
\end{itemize}

\subsection{Decoupled Semantic and Geometric Pathways}
To fully unleash the potential of the decoupled priors from our foundation encoder, we introduce decoupled semantic and geometric pathways. Each pathway is enhanced by a specialised module that is specifically designed to address its unique challenges and maximise its contributions.

\subsubsection{Geometry-Aware Context Adapter.}
While VFM features exhibit impressive semantic capabilities, their 2D-centric pre-training limits understanding of 3D geometry~\cite{el2024probing}, leading to geometric inconsistencies when applied to 3D perception tasks. One possible solution is to fine-tune these models using 3D data~\cite{yue2024improving}; however, this approach can involve significant computational costs. To address this issue without expensive fine-tuning, we introduce the Geometry-Aware Context Adapter (GCA), a lightweight module designed to incorporate 3D structural awareness into the existing VFM priors.

The GCA begins by constructing a comprehensive geometry prior, $\mathbf{M}^{g}$, from the depth map $\mathbf{Z}$. This prior matrix fuses two distinct relationships between all pixel pairs: a depth relationship matrix, $\mathbf{M}^{d}$, which encodes their metric distance in 3D space, and a spatial distance matrix, $\mathbf{M}^{s}$, which represents their Manhattan distance on the 2D image plane. These matrices are balanced by a learnable parameter $\alpha \in [0, 1]$:
\begin{equation}
    \begin{split}
    &\mathbf{M}^{d}_{ij,i'j'}=|\mathbf{Z}_{ij}-\mathbf{Z}_{i'j'}|\\
    &\mathbf{M}^{s}_{ij,i'j'}=|i-i'|+ |j-j'| \\
    &\mathbf{M}^{g}=\alpha \mathbf{M}^{d} + (1-\alpha) \mathbf{M}^{s}\quad .
    \end{split}
    \label{eq:geometry_prior}
\end{equation}
where matrices are of shape $HW \times HW$ ($H$ and $W$ denote the height and width of the feature map). This matrix $\mathbf{M}^g$ serves as a structural bias that is integrated directly into the geometry self-attention mechanism. 

Specifically, the standard attention map is multiplied element-wise by a term derived from $\mathbf{M}^g$. This operation effectively directs the attention to concentrate on geometrically coherent pixel relationships. It achieves this by reducing the connections between pixels that are close in 2D but distant in 3D. The operation can be expressed as:
\begin{equation}
    \operatorname{GeoAttn}(\mathbf{Q},\mathbf{K},\mathbf{V},\mathbf{M}^g)=(\operatorname{Softmax}(\mathbf{QK}^T)\odot \beta^{\mathbf{M}^g})\mathbf{V}\quad .
    \label{eq:gsa}
\end{equation}

where $\odot$ denotes element-wise multiplication and $\beta \in (0,1)$ is a learnable decay rate. To achieve both representational diversity and computational efficiency, we follow the approach outlined in~\cite{yin2025dformerv2} by assigning distinct decay rates to different attention heads. Additionally, we implement the attention mechanism using an axes decomposition strategy. By incorporating explicit 3D structure into the attention modulation, GCA enhances the 2D VFM features $\mathbf{F}^{2D}$, improving their geometric consistency. These refined features then serve as enhanced context features $\mathbf{C}$ for the process of lifting into the 3D voxel space.

\begin{figure}[t!]
  \centering
    \includegraphics[width=0.98\linewidth]{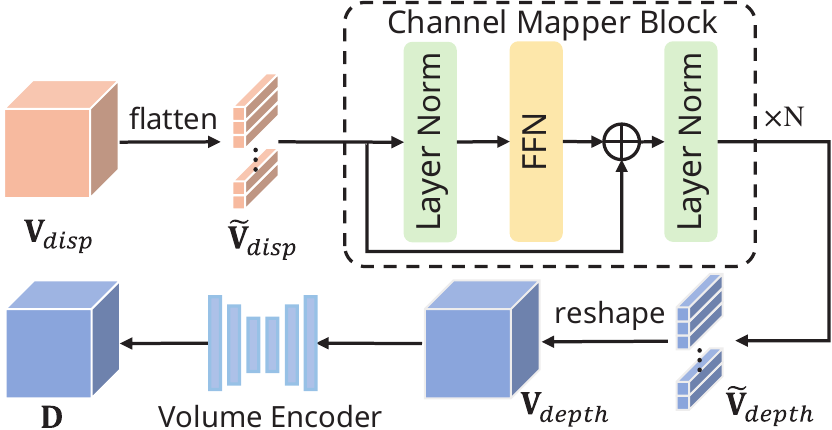}
    \caption{Illustration of the DDVM module, which transforms disparity cost volume to depth distribution.}
    \label{fig:DDVM}
\end{figure}

\subsubsection{Disparity-to-Depth Volume Mapping.}
The stereo cost volume, $\mathbf{V}_{disp}$, contains rich, probabilistic information about geometric correspondences, offering a far more detailed representation of 3D structure than a single depth map. Unlocking this potential is a core objective of our framework. 
However, its disparity-aligned structure is incompatible with the depth-binned format required by the LSS pipeline.
The conventional solution is to first collapse the cost volume into a deterministic depth map and then build a one-hot depth distribution. This process acts as a significant information bottleneck by irretrievably discarding crucial details about matching ambiguities and uncertainty. 

To address the issue of information loss, we propose the Disparity-to-Depth Volume Mapping (DDVM) module. This lightweight network learns a direct, non-linear mapping from a disparity-based volume to a depth-based volume. As illustrated in Fig.~\ref{fig:DDVM}, the module transforms features from the disparity dimension into the depth dimension.

Given the cost volume features $\mathbf{V}_{disp}\in\mathbb{R}^{D_{disp}\times H\times W}$, the spatial dimensions are flattened to reshape the representation into $\mathbf{\tilde{V}}_{disp}\in\mathbb{R}^{D_{disp}\times HW}$. While $\mathbb{R}^{D_{disp}\times H\times W}$ preserves the 2D spatial layout, $\mathbb{R}^{D_{disp}\times HW}$ denotes its flattened form. A non-linear mapping function, $f(\cdot)$, transforms the feature channels directly from disparity to depth:

\begin{equation}
    \mathbf{\tilde{V}}_{depth}=f(\mathbf{\tilde{V}}_{disp})\quad .
\end{equation}
where the output $\mathbf{\tilde{V}}_{depth}$ has the shape $\mathbb{R}^{D_{depth}\times HW}$. The mapping function $f(\cdot)$ comprises stacked channel-mapper blocks, each consisting of a feed-forward network (FFN), layer normalisation, and a residual connection. This design enables the network to learn complex, non-linear correspondences between disparity and depth channels.

To restore spatial structure, $\mathbf{\tilde{V}}_{depth}$ is reshaped into 3D volumes $\mathbf{V}_{depth}\in\mathbb{R}^{D_{depth} \times H\times W}$. A refinement module then employs a shallow 3D CNN encoder $\phi(\cdot)$ to enhance spatial coherence and smoothness, followed by a softmax operation along the depth dimension to produce the probabilistic depth distribution, $\mathbf{D} \in \mathbb{R}^{D_{depth} \times H\times W}$:
\begin{equation}
\mathbf{D}=\operatorname{Softmax}(\phi(\mathbf{V}_{depth}))\quad .
\end{equation}

Unlike methods that depend on explicit analytical conversion, our learned mapping is more flexible and robust. It retains the detailed probabilistic characteristics of the original cost volume, offering a much richer geometric prior for the subsequent view transformation.

\subsection{Hybrid View Transformation}
We follow~\cite{yu2024CGFormer} for 2D-to-3D lifting, which integrates an LSS-based voxel initialisation with transformer-based voxel refinement. In the LSS view transformation, the context features $\mathbf{C}\in\mathbb{R}^{C \times H \times W}$ and the categorical depth distributions $\mathbf{D} \in \mathbb{R}^{D\times H \times W}$ are combined via the outer product $\mathbf{C} \otimes \mathbf{D}$ to construct the 3D frustum representation $\mathbf{G} \in \mathbb{R}^{C \times D \times H \times W}$. After this, voxel pooling~\cite{philion2020lss} is applied to splat frustum features $\mathbf{G}$ into a 3D voxel grid, resulting in the 3D feature representation $\mathbf{F}_{lss} \in \mathbb{R}^{C \times X \times Y \times Z}$, where $(X,\ Y,\ Z)$ denotes the spatial resolution of the voxel grid. In the voxel transformer, the LSS volume and a depth map are utilised to generate query proposals $\mathbf{F}_q$, where the depth map first identifies a sparse set of occupied voxel locations, and the LSS volume provides their initial features. Deformable cross-attention is employed to gather relevant 3D frustum features for each proposal, followed by deformable self-attention among voxels to produce the final 3D feature representation $\mathbf{F}_{vt}\in \mathbb{R}^{C \times X \times Y \times Z}$.

Inspired by~\cite{li2023fbbev, li2024dualbev}, we hypothesise that the two 3D feature volumes contain complementary information: $\mathbf{F}_{lss}$ preserves precise geometric details for near-field objects, while $\mathbf{F}_{vt}$ captures superior contextual information for distant and occluded regions. However, previous methods, such as those proposed by~\cite{yu2024CGFormer}, often overlook the fusion of these features, treating $\mathbf{F}_{lss}$  as a disposable intermediate representation and thereby risking the loss of valuable information. Effectively fusing these complementary representations is a critical challenge that is often neglected, and we will address this issue next.

\subsubsection{Axis-Aware Fusion.}
Our initial fusion exploration started with a vanilla 3D channel attention method inspired by 2D techniques~\cite{dai2021aff}. Although this method demonstrated some performance improvements, we believe its effectiveness is constrained by the isotropic nature of its fusion mechanisms. Specifically, the 3D global pooling operation compresses the entire volume into a single descriptor, which results in the loss of crucial directional cues.

To enable spatially aware 3D feature fusion, we introduce the Axis-Aware Fusion (AAF) module. The core concept behind AAF is to independently capture and merge features based on contextual information along each of the three primary axes ($X,\ Y,\ Z$). To accomplish this, the AAF module utilises three parallel fusion units, with each unit dedicated to one axis and operating on the orthogonal plane to extract axis-specific context. For example, the unit for the $Z$-axis works on the $XY$-plane. 

\begin{figure}[t!]
  \centering
    \includegraphics[width=0.98\linewidth]{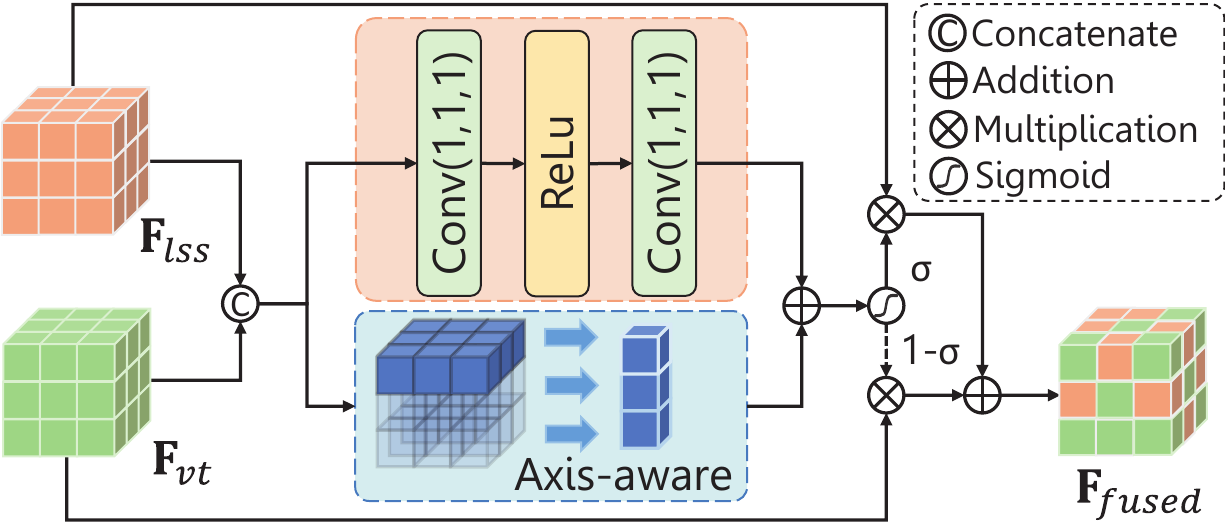}
    \caption{
    Illustration of the fusion unit for a single axis (e.g., the $Z$-axis) within the proposed AAF module.}
    \label{fig:AAF}
\end{figure}

As illustrated in Fig.~\ref{fig:AAF}, within each unit, a dynamic, axis-specific attention map, denoted as $\sigma_d$, is created by synthesising both local and global information. Specifically, the input features ($\mathbf{F}_{lss}$ and $\mathbf{F}_{vt}$), are processed through two simultaneous pathways: the local pathway employs 3D convolutions to capture fine-grained spatial details, while the global pathway utilises our anisotropic pooling strategy (pooling along the two axes of the plane) to extract a global context feature specific to the orthogonal axis.

The features from both pathways are then fused to generate the attention map $\sigma_d$. The final fused feature for the entire module, $\mathbf{F}_{fused}$, is obtained by summing the outputs from the three parallel, axis-specific fusion units:
\begin{equation}
    \mathbf{F}_{fused}=\sum_{d \in {XY,\ XZ,\ YZ}} (\sigma_{d}\mathbf{F}_{lss} + (1-\sigma_{d})\mathbf{F}_{vt})\quad .
    \label{eq:aaf}
\end{equation}

Eq.~\ref{eq:aaf} leverages both local and global information specific to each axis. As a result, the AAF module effectively merges complementary features in a highly adaptive and spatially aware way. The fusion operation produces a unified 3D feature representation, enhanced by a thorough understanding of the scene's structure along all three orthogonal axes.

\begin{table*}[ht]
\centering
\small
\setlength{\tabcolsep}{1.12mm}
\begin{tabular}{l|cc|ccccccccccccccccccc}
    \toprule
    Method
    &IoU
    &mIoU
    & \rotatebox{90}{road} 
    & \rotatebox{90}{sidewalk}
    & \rotatebox{90}{parking}
    & \rotatebox{90}{other-grnd.}
    & \rotatebox{90}{building}
    & \rotatebox{90}{car}
    & \rotatebox{90}{truck}
    & \rotatebox{90}{bicycle}
    & \rotatebox{90}{motorcycle}
    & \rotatebox{90}{other-veh.}
    & \rotatebox{90}{vegetation}
    & \rotatebox{90}{trunk}
    & \rotatebox{90}{terrain}
    & \rotatebox{90}{person}
    & \rotatebox{90}{bicyclist}
    & \rotatebox{90}{motorcyclist} 
    &  \rotatebox{90}{fence}
    & \rotatebox{90}{pole}
    & \rotatebox{90}{traf.-sign}
    \\
    \midrule
    MonoScene &34.16	&11.08	&54.7	&27.1	&24.8	&5.7	&14.4	&18.8	&3.3	&0.5	&0.7	&4.4	&14.9	&2.4	&19.5	&1.0	&1.4	&0.4	&11.1	&3.3	&2.1 \\
    
    TPVFormer &34.25	&11.26	&55.1	&27.2	&27.4	&6.5	&14.8	&19.2	&3.7	&1.0	&0.5	&2.3	&13.9	&2.6	&20.4	&1.1	&2.4	&0.3	&11.0	&2.9	&1.5 \\
    

    VoxFormer	&43.21	&13.41	&54.1	&26.9	&25.1	&7.3	&23.5	&21.7	&3.6	&1.9	&1.6	&4.1	&24.4	&8.1	&24.2	&1.6	&1.1	&0.0	&13.1	&6.6	&5.7 \\
    

    OccFormer	&34.53	&12.32	&55.9	&30.3	&31.5	&6.5	&15.7	&21.6	&1.2	&1.5	&1.7	&3.2	&16.8	&3.9	&21.3	&2.2	&1.1	&0.2	&11.9	&3.8	&3.7 \\

    



    

    Symphonies &42.19	&15.04	&58.4	&29.3	&26.9	&11.7	&24.7	&23.6	&3.2	&3.6	&2.6	&5.6	&24.2	&10.0	&23.1	&\underline{3.2}	&1.9	&\underline{2.0}	&16.1	&7.7	&8.0 \\

    BRGScene &43.34	&15.36	&61.9	&31.2	&30.7	&10.7	&24.2	&22.8	&2.8	&3.4	&2.4	&6.1	&23.8	&8.4	&27.0	&2.9	&2.2	&0.5	&16.5	&7.0	&7.2 \\
    
    HTCL	&44.23	&17.09	&64.4	&34.8	&33.8	&12.4	&25.9	&27.3	&5.7	&1.8	&2.2	&5.4	&25.3	&10.8	&31.2	&1.1	&3.1	&0.9	&21.1	&9.0	&8.3 \\
    
    CGFormer	&44.41	&16.63	&64.3	&34.2	&34.1	&12.1	&25.8	&26.1	&4.3	&3.7	&1.3	&2.7	&24.5	&11.2	&29.3	&1.7	&3.6	&0.4	&18.7	&8.7	&9.3 \\

    VLScene	&45.14	&17.52	&64.7	&34.7	&32.4	&\underline{13.1}	&27.3	&26.1	&\underline{6.5}	&4.2	&3.8	&\underline{8.3}	&26.4	&10.0	&29.4	&2.8	&\underline{5.1}	&0.9	&20.0	&8.9	&8.4 \\
    
    ScanSSC	&44.54	&17.40	&\textbf{66.2}	&35.9	&\underline{35.1}	&12.5	&25.3	&27.1	&3.5	&3.5	&3.2	&6.1	&25.2	&11.0	&30.6	&1.8	&\textbf{5.3}	&0.7	&20.5	&8.4	&8.9 \\

    SOAP	&\underline{46.09}	&\underline{19.09}	&63.6	&\underline{36.2}	&\textbf{36.8}	&\textbf{17.2}	&\underline{28.7}	&\textbf{28.9}	&3.3	&\underline{5.0}	&\underline{5.3}	&7.0	&\textbf{29.8}	&\textbf{15.1}	&\underline{32.3}	&\textbf{3.7}	&2.3	&0.2	&\underline{22.5}	&\textbf{11.7}	&\textbf{13.1} \\ \hline

    Ours    &\textbf{48.12}	&\textbf{19.32}	&\underline{65.7}	&\textbf{36.3}	&33.5	&11.6	&\textbf{30.5}	&\underline{28.2}	&\textbf{9.0}	&\textbf{5.6}	&\textbf{6.9}	&\textbf{10.0}	&\underline{28.6}	&\underline{13.7}	&\textbf{32.8}	&2.8	&4.7	&\textbf{2.5}	&\textbf{23.1}	&\underline{9.8}	&\underline{11.8} \\
    \bottomrule
\end{tabular}
\caption{Quantitative results on SemanticKITTI test set. Best and second-best results are in \textbf{bold} and \underline{underlined}.}
\label{tab:sem_kitti_test}
\end{table*}
\begin{table*}[ht]
\centering
\small
\setlength{\tabcolsep}{1.29mm}
\begin{tabular}{l|cc|ccccccccccccccccccc}
    \toprule
    Method
    &IoU
    &mIoU
    & \rotatebox{90}{car}
    & \rotatebox{90}{bicycle} 
    & \rotatebox{90}{motorcycle}  
    & \rotatebox{90}{truck}
    & \rotatebox{90}{other-veh.}
    & \rotatebox{90}{person}
    & \rotatebox{90}{road}
    & \rotatebox{90}{parking}
    & \rotatebox{90}{sidewalk}
    & \rotatebox{90}{other-grnd.}
    & \rotatebox{90}{building}
    & \rotatebox{90}{fence}
    & \rotatebox{90}{vegetation}
    & \rotatebox{90}{terrain}
    & \rotatebox{90}{pole}
    & \rotatebox{90}{traf.-sign}
    &  \rotatebox{90}{other-struct.}
    & \rotatebox{90}{other-obj.}
    \\
    \midrule
    MonoScene	&37.87	&12.31	&19.3 	&0.4 	&0.6 	&8.0 	&2.0 	&0.9 	&48.4 	&11.4 	&28.1 	&3.3 	&32.9 	&3.5 	&26.2 	&16.8 	&6.9 	&5.7 	&4.2 	&3.1 \\

    VoxFormer	&38.76	&11.91	&17.8 	&1.2 	&0.9 	&4.6 	&2.1 	&1.6 	&47.0 	&9.7 	&27.2 	&2.9 	&31.2 	&5.0 	&29.0 	&14.7 	&6.5 	&6.9 	&3.8 	&2.4 \\
    
    TPVFormer	&40.22	&13.64	&21.6 	&1.1 	&1.4 	&8.1 	&2.6 	&2.4 	&53.0 	&12.0 	&31.1 	&3.8 	&34.8 	&4.8 	&30.1 	&17.5 	&7.5 	&5.9 	&5.5 	&2.7 \\
    
    OccFormer	&40.27	&13.81	&22.6 	&0.7 	&0.3 	&9.9 	&3.8 	&2.8 	&54.3 	&13.4 	&31.5 	&3.6 	&36.4 	&4.8 	&31.0 	&19.5 	&7.8 	&8.5 	&7.0 	&4.6 \\
    
    Symphonies	&44.12	&18.58	&\underline{30.0}	&1.9	&5.9	&\textbf{25.1}	&\underline{12.1}	&\textbf{8.2}	&54.9	&13.8	&32.8	&\textbf{6.9}	&35.1	&8.6	&38.3	&11.5	&14.0	&9.6	&\textbf{14.4}	&\textbf{11.3} \\
    
    
    CGFormer	&48.07	&20.05	&29.9	&3.4	&4.0	&17.6	&6.8	&6.6	&\textbf{63.9}	&17.2	&\textbf{40.7}	&5.5	&42.7	&8.2	&38.8	&\underline{24.9}	&16.2	&17.5	&10.2	&6.8 \\
    
    VLScene	&46.08	&19.10	&29.0	&4.7	&7.7	&18.3	&7.6	&\underline{7.4}	&60.1	&17.4	&39.0	&\underline{6.0}	&42.1	&9.6	&36.5	&24.8	&\textbf{17.0}	&18.8	&10.5	&6.5 \\

    ScanSSC	&\underline{48.29}	&20.14	&29.9	&3.8	&4.3	&14.3	&9.1	&6.7	&62.2	&\textbf{18.2}	&40.2	&5.2	&42.7	&8.8	&38.8	&\textbf{25.5}	&16.6	&19.1	&10.3	&6.9 \\

    SOAP	&48.12	&\underline{20.92}	&29.9	&\textbf{5.6}	&\underline{7.8}	&14.4	&7.6	&6.1	&60.9	&17.4	&\underline{40.3}	&5.4	&\textbf{45.3}	&\textbf{10.6}	&\textbf{40.5}	&24.8	&\underline{16.8}	&\textbf{21.0}	&\underline{12.6}	&9.9 \\ \hline

    Ours    &\textbf{48.61} 	&\textbf{21.78} 	&\textbf{30.5} 	&\underline{4.9} 	&\textbf{8.4} 	&\underline{24.9} 	&\textbf{13.0} 	&7.2 	&\underline{63.6} 	&\underline{17.8} 	&39.8 	&5.7 	&\underline{43.6} 	&\underline{10.5} 	&\underline{39.1} 	&24.8 	&16.2 	&\underline{20.3} 	&11.5 	&\underline{10.2} \\
    
    \bottomrule
\end{tabular}

\caption{Quantitative results on SSCBench-KITTI-360 test set. Best and second-best results are in \textbf{bold} and \underline{underlined}.
}
\label{tab:kitti_360_test}
\end{table*}

\subsection{Training Loss}
To train the FoundationSSC, we follow MonoScene~\cite{cao2022monoscene}, utilise affinity losses $\mathcal{L}_{scal}^{geo}$ and $\mathcal{L}_{scal}^{sem}$ to optimise the scene-wise and class-wise metrics, in conjunction with the cross-entropy loss $\mathcal{L}_{ce}$ weighted by class frequencies. Additionally, we employ two auxiliary losses: a 2D semantic segmentation loss $\mathcal{L}_{s}$ for feature enrichment and a depth loss $\mathcal{L}_{d}$~\cite{li2024bevformer} for geometric supervision. The total loss is a weighted sum of these components:
\begin{equation}
	\mathcal{L} = \lambda_{d}\mathcal{L}_{d} + \lambda_{s}\mathcal{L}_{s}+ \mathcal{L}_{ce} + \mathcal{L}_{scal}^{geo} + \mathcal{L}_{scal}^{sem}\quad .
\end{equation}
where $\lambda_{d}$ and $\lambda_{s}$ are the weights for the auxiliary losses. Following preliminary experiments, we set $\lambda_{d}=0.001$ and $\lambda_{s}=1$, respectively.

\section{Experiment}
\label{sec:exp}
We evaluate our FoundationSSC on SemanticKITTI~\cite{behley2019semantickitti} and SSCBench-KITTI-360~\cite{li2024sscbench}. Following previous methods~\cite{cao2022monoscene, yu2024CGFormer}, we report the IoU and mIoU as evaluation metrics. Refer to the supplementary material for more information about datasets, metrics, and implementation details.

\subsection{Quantitative Results}
We present the quantitative results on SemanticKITTI hidden test set. As summarised in Table~\ref{tab:sem_kitti_test}, FoundationSSC achieves state-of-the-art performance with a mIoU of 19.32 and an IoU of 48.12, demonstrating a +2.69 increase in mIoU and a +3.71 increase in IoU compared to the CGFormer baseline~\cite{yu2024CGFormer}. The simultaneous and substantial gains across both metrics provide direct evidence that the dual-decoupling design effectively balances the inherent trade-off of coupled architectures. Furthermore, despite relying solely on stereo images as input, FoundationSSC outperforms all temporal methods, including HTCL~\cite{li2024htcl} and SOAP~\cite{lee2025soap}. On a per-class level, our model achieves the best or second-best performance in most categories. Benefiting from the strong feature priors provided by the foundation encoder, our model demonstrates particularly robust results in long-tail classes, such as bicycle and motorcycle.

We further report results on SSCBench-KITTI-360 in Table~\ref{tab:kitti_360_test}, where FoundationSSC once again achieves top performance with 48.61 IoU and a 21.78 mIoU. The consistent superiority across two benchmarks underscores the robustness and generalizability of our proposed framework.

\subsection{Ablation Study}
We conduct ablation studies on SemanticKITTI validation set to evaluate the effectiveness of each component.

\subsubsection{Ablation on the Architectural Components.}
Table~\ref{tab:arch_ablation} analyses the contribution of each component in FoundationSSC. Our baseline is a simplified CGFormer variant that excludes the TPV branch. 
First, we assess the impact of source-level decoupling (a). Replacing the baseline's encoder with our Foundation Encoder (FE) results in substantial gains across both metrics (+2.06 mIoU, +1.33 IoU), validating the effectiveness of using high-quality, disentangled priors.
Next, we ablate the components of our pathway-level decoupling based on the FE-only model. The GCA module (b) enhances both semantic and geometric features (+0.36 mIoU, +0.56 IoU) by injecting 3D awareness. The DDVM module (c) provides an even larger, geometry-focused boost (+0.52 mIoU, +1.32 IoU) by preserving probabilistic information. Moreover, combining both modules (d) results in a synergistic improvement (+0.97 mIoU, +1.23 IoU) over the FE-only model, demonstrating that our specialised pathways work together effectively. 
Lastly, adding the AAF module (e) to the full pathway model (d) contributes to further performance enhancements (+0.80 mIoU, +0.07 IoU), verifying our anisotropic fusion strategy. In total, our whole model demonstrates remarkable gains of +3.83 mIoU and +2.63 IoU over the baseline, confirming that each component of our dual-decoupling design effectively contributes to both semantic and geometric understanding.

\begin{table}[t]
\centering
\small
\setlength{\tabcolsep}{2.3mm}
\begin{tabular}{c|cccc|cc}
    \toprule
     \textbf{Method} &{FE} &{GCA} &{DDVM} &{AAF} &\textbf{IoU} &\textbf{mIoU} \\
     \midrule
     Baseline &{} &{} &{} &{} &{45.28} &{16.53} \\
     (a) &{\checkmark} &{} &{} &{} &{46.61} &{18.59}  \\
     (b) &{\checkmark} &{\checkmark} &{} &{} &{47.17} &{18.95}  \\
     (c) &{\checkmark} &{} &{\checkmark} &{} &{47.93} &{19.11}  \\
     (d) &{\checkmark} &{\checkmark} &{\checkmark} &{} &{47.84} &{19.56} \\
     (e) &{\checkmark} &{\checkmark} &{\checkmark} &{\checkmark} &{47.91} &{20.36}  \\
    \bottomrule
\end{tabular}
  \caption{Ablation study for architectural components.}
  \label{tab:arch_ablation}
\end{table}

\begin{table}[t]
  \centering
  \small
  \setlength{\tabcolsep}{3.6mm}
  \begin{tabular}{l|cc|cc}
    \toprule
    Backbone Networks &IoU    &mIoU   &Parameters \\
    \midrule
    EfficientNetB7    &44.55  &15.20  &66.35M      \\
    DINOv2-S          &45.33  &15.52  &22.06M      \\
    DepthAnythingV2-S &45.31  &15.59  &24.79M      \\
    DINOv2-L          &45.88  &\textbf{17.79}  &304.37M      \\
    DepthAnythingV2-L &\textbf{46.19}  &17.01  &335.32M     \\
    \bottomrule
  \end{tabular}
  \caption{Ablation study on the image backbone.}
  \label{tab:diff_backbone}
\end{table}

\begin{figure*}[t!]
    \includegraphics[width=\textwidth]{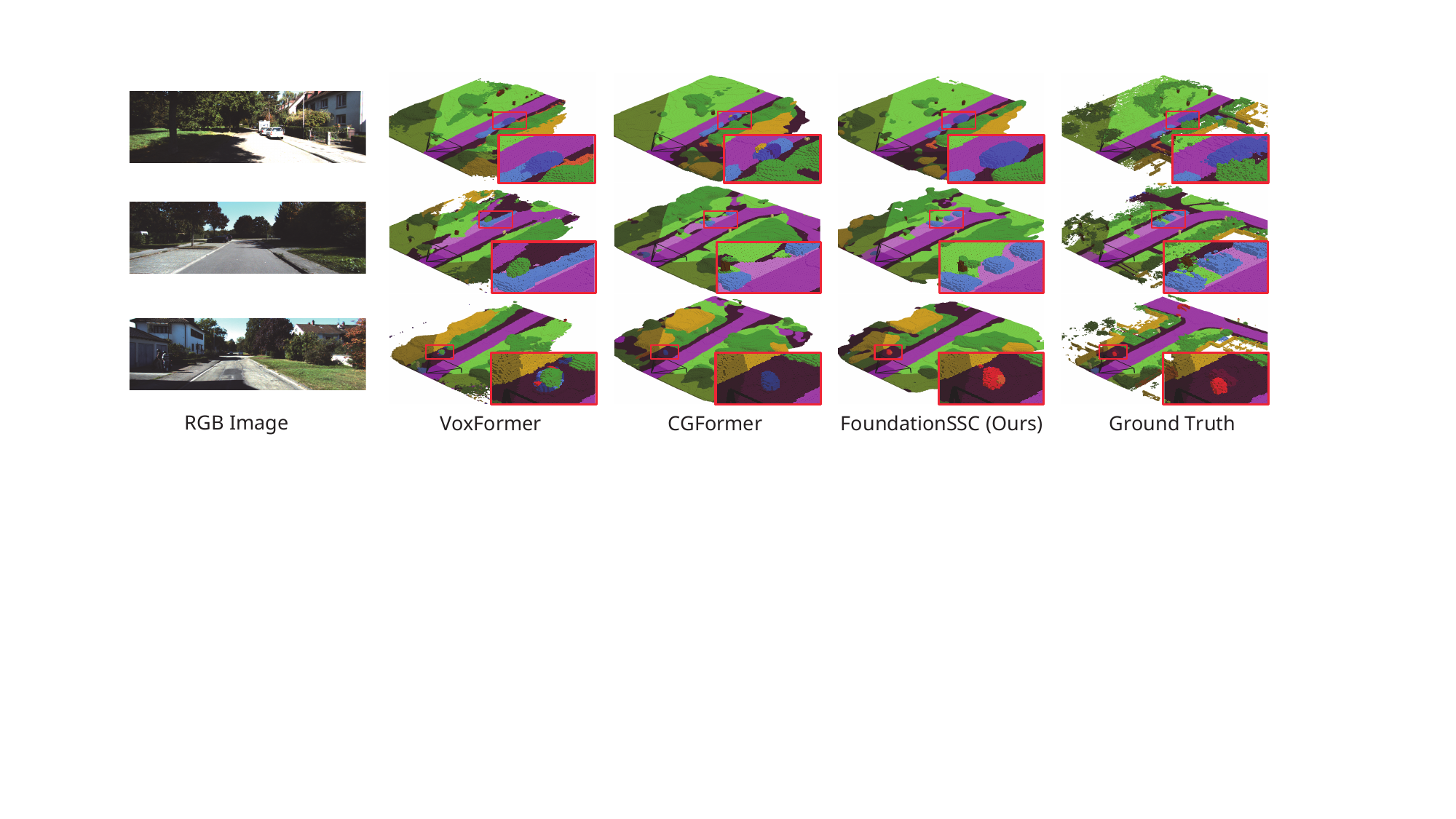}
    \caption{Qualitative visualisation results on SemanticKITTI validation set. Our FoundationSSC predicts objects with sharper geometric boundaries and more accurate semantic labels, outperforming competing works.}
    \label{fig:vis}
\end{figure*}

\subsubsection{Ablation Study for Foundation Encoder.}
To evaluate the impact of different VFMs, we conduct an ablation study where each model derives context features and monocular depth distributions from shared internal representations, enabling a fair comparison of semantic-geometric trade-offs. As shown in Table~\ref{tab:diff_backbone}, all VFM variants outperform the EfficientNet-B7~\cite{tan2019efficientnet} baseline, demonstrating superior representational capacity. Among them, DINOv2-L~\cite{oquab2024dinov2} achieves the highest mIoU, while the depth-specialised DepthAnythingV2-L~\cite{yang2024depthanythingv2} attains higher IoU at the expense of a slightly lower mIoU. The performance contrast highlights the specialisation-generalisation trade-off and supports a decoupled design philosophy. Although pairing separate backbones (e.g. DINOv2 for semantics and another for geometry) is feasible, such configurations increase architecture complexity. Instead, we adopt the frozen DepthAnythingV2-L within the unified FoundationStereo encoder, ensuring both simplicity and efficiency.

\subsubsection{Ablation Study for DDVM.}
We validate our DDVM module against various depth generation strategies in Table~\ref{tab:ddvm}. The CGFormer~\cite{yu2024CGFormer} depth refinement module, which fuses monocular distribution with a pre-processed stereo depth map using neighbour attention, achieves 19.83 mIoU and 47.87 IoU. A key limitation of such a method is its reliance on information-lossy depth maps, which discard valuable probabilistic cues from stereo matching.
We then explore the direct use of the disparity cost volume. A naive approach using analytical resampling (AR) degrades performance to 19.59 mIoU, confirming that merely accessing the cost volume is insufficient. In contrast, our DDVM, by replacing rigid resampling with a learnable mapping, achieves a state-of-the-art mIoU of 20.36. The performance gain demonstrates DDVM's superior ability to preserve and leverage fine-grained geometric cues for SSC.
 
\begin{table}[t]
  \centering
  \small
\setlength{\tabcolsep}{6.5mm}
  \begin{tabular}{l|ccc}
    \toprule
    {\textbf{Setting}} & {\textbf{IoU}} & {\textbf{mIoU}} \\
    \midrule
    Depth Refinement           & 47.87   & 19.83    \\
    Cost Volume + AR          &47.76    &19.59  \\
    Cost Volume + DDVM         &\textbf{47.91}      &\textbf{20.36}    \\
    \bottomrule
  \end{tabular}
  \caption{Ablation study on the depth generation strategies.}
  \label{tab:ddvm}
\end{table}

\begin{table}[t]
  \centering
  \small
 \setlength{\tabcolsep}{9.5mm}
  \begin{tabular}{l|ccc}
    \toprule
    {\textbf{Setting}} & {\textbf{IoU}} & {\textbf{mIoU}} \\
    \midrule
    w/o AAF     &47.84    &19.56   \\
    3D CA        &\textbf{48.25}   & 20.08   \\
    AAF         &47.91  &{\textbf{20.36}}  \\
    \bottomrule
  \end{tabular}
  \caption{Ablation study on the 3D feature fusion strategy.}
  \label{tab:aaf}
\end{table}

\subsubsection{Ablation Study for AAF.}
Table~\ref{tab:aaf} compares the AAF module against alternative 3D feature fusion strategies. We establish a baseline by using only the refined volume $\mathbf{F}_{vt}$, mirroring prior work, which yields 19.56 mIoU. A fusion approach using a vanilla 3D channel attention (3D-CA) module improves performance to 20.08 mIoU. However, the effectiveness of 3D-CA is constrained by an isotropic global pooling, a design that neglects directional cues critical for 3D understanding. In contrast, our AAF module is explicitly designed to be anisotropic. The module's axis-specific fusion mechanism achieves the highest mIoU of 20.36. The substantial performance gain over the 3D-CA provides compelling evidence that modelling the anisotropic nature of 3D scenes is crucial for effective feature fusion.

\subsection{Qualitative Results} 
Fig.~\ref{fig:vis} presents a qualitative comparison on SemanticKITTI validation set, showcasing the results of our FoundationSSC, VoxFormer~\cite{li2023voxformer}, CGFormer~\cite{yu2024CGFormer}, and the ground truth. The visual results highlight the dual advantages of our framework in both geometry and semantics. Our geometric superiority is evident with distant, partially occluded objects. While competing methods produce blurry, merged voxels, FoundationSSC renders them with crisp, well-defined boundaries. Our semantic advantage is showcased in identifying smaller or less common classes. While competing methods may misclassify objects, such as pedestrians, FoundationSSC accurately predicts their semantic labels. These visual advantages are a direct outcome of our dual-decoupling strategy, enabling the framework to leverage and process robust, disentangled priors effectively.

\section{Conclusion}
\label{sec:conclusion}
In this paper, we introduce FoundationSSC, a novel neural network designed to resolve the semantic-geometric conflict in camera-based SSC through a dual-decoupling approach. FoundationSSC features an integrated foundation encoder that provides robust and disentangled semantic and geometric priors. These priors are refined through specialised decoupled pathways and then combined using our innovative Axis-Aware Fusion module. The effectiveness of this strategy is demonstrated by achieving state-of-the-art results on SemanticKITTI and SSCBench-KITTI-360 datasets. The substantial improvements in both geometric (IoU) and semantic (mIoU) metrics confirm that our framework can properly balance the long-standing performance trade-off.

\section{Appendix/Supplemental Material}
In the appendix, we mainly provide implementation details and more experiment results.

\subsection{Experimental Setup}
\noindent\textbf{Datasets.} We evaluate our FoundationSSC on two widely-used datasets: SemanticKITTI~\cite{behley2019semantickitti} and SSCBench-KITTI-360~\cite{li2024sscbench}. SemanticKITTI provides RGB images of size $1226 \times 370$ and includes 20 classes (19 semantic classes and one free class). The dataset comprises 10 training sequences, one validation sequence, and 11 testing sequences. SSCBench-KITTI-360~\cite{li2024sscbench} offers input RGB images at $1408 \times 376$ resolution. It contains 19 classes (18 semantic classes and one free class) with seven training sequences, one validation sequence, and one testing sequence. All evaluations on these two benchmarks focus on a defined spatial area: $51.2m$ in front of the vehicle, $25.6m$ to both the left and right, and $6.4m$ above the vehicle. Voxelizing this region results in a set of 3D voxel grids with a resolution of $256 \times 256 \times 32$, where each voxel measures $0.2m$ on each side. 

\noindent\textbf{Metrics.} Following previous methods~\cite{cao2022monoscene, li2023voxformer, yu2024CGFormer}, we report the Intersection over Union (IoU) and mean IoU (mIoU) metrics for occupied voxel completion and voxel-wise semantic completion evaluation, respectively. These two metrics offer a comprehensive view of the model's ability to reconstruct the scene in terms of both geometry and semantics.

\subsection{Implementation Details}
\noindent\textbf{Image Encoder.} We adopt the frozen FoundationStereo~\cite{wen2025foundationstereo} as the image encoder and leverage its backbone, DepthAnythingV2~\cite{yang2024depthanythingv2}, to extract multi-level image features. Since a ViT-based backbone is non-hierarchical, as in~\cite{li2022vitdet}, we utilise the last and strongest feature map to produce multi-scale feature maps. We then align them in both spatial and channel dimensions, and finally concatenate them along the channel dimensions into a stacked feature tensor. Note that DepthAnythingV2 originally outputs feature maps at a $1/14$ resolution. For easy access to subsequent feature utilisation, we interpolate the multi-level feature maps to $1/16$. After passing them through the FPN and stacking the multi-scale feature maps, the final image feature tensor $\mathbf{F}^{2D}$ is at a $1/8$ resolution.

The input resolution is $384 \times 1280$ on SemanticKITTI and $384 \times 1408$ on SSCBench-KITTI-360 to match the patch size of the FoundationStereo encoder.

\noindent\textbf{Training Setup.} We train FoundationSSC for 25 epochs on 4 NVIDIA 4090 GPUs, with a batch size of 4. It approximately consumes 22 GB of GPU memory on each GPU during the training phase. We employ the AdamW~\cite{loshchilov2018adamw} optimizer with $\beta_{1} = 0.9$, $\beta_{2} =0.99$ and set the maximum learning rate to $3\times 10^{-4}$. The cosine annealing learning rate strategy is adopted for the learning rate decay, where the cosine warmup strategy is applied for the first $5\%$ iterations.

\subsection{Quantitative Results}
To provide a more thorough comparison, we report additional quantitative results on SemanticKITTI validation set in Table~\ref{tab:sem_kitti_val}. The state-of-the-art performance further validates the effectiveness of our approach in advancing 3D scene perception.

\subsection{Qualitative Results}
\begin{figure*}[t!]
    \includegraphics[width=0.98\textwidth]{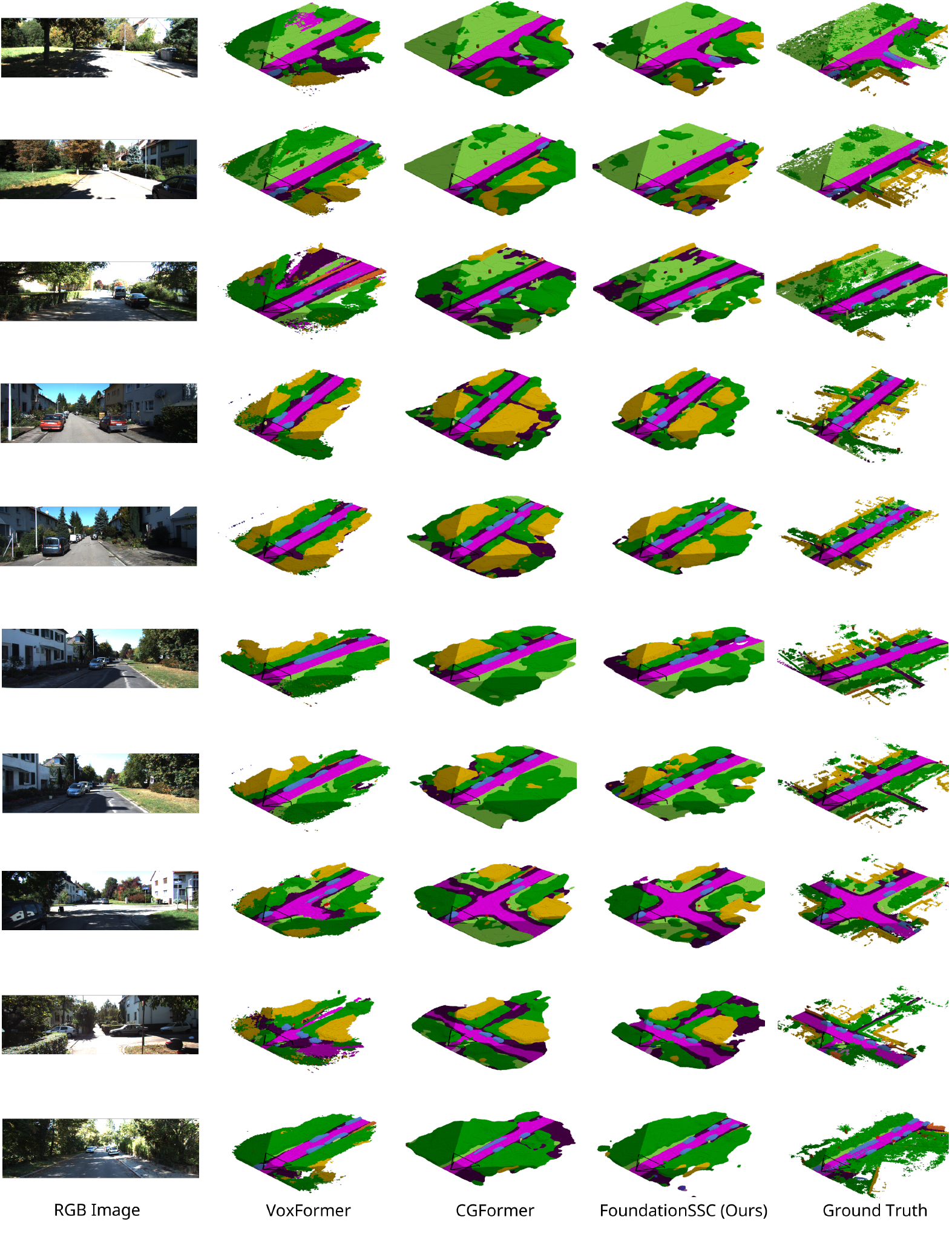}
    \caption{More qualitative comparison results on SemanticKITTI validation set.}
    \label{fig:vis_sup}
\end{figure*}

We report additional visualization results on SemanticKITTI validation set in Fig.~\ref{fig:vis_sup}.

\begin{figure}[t!]
    \centering
    \includegraphics[width=0.98\linewidth]{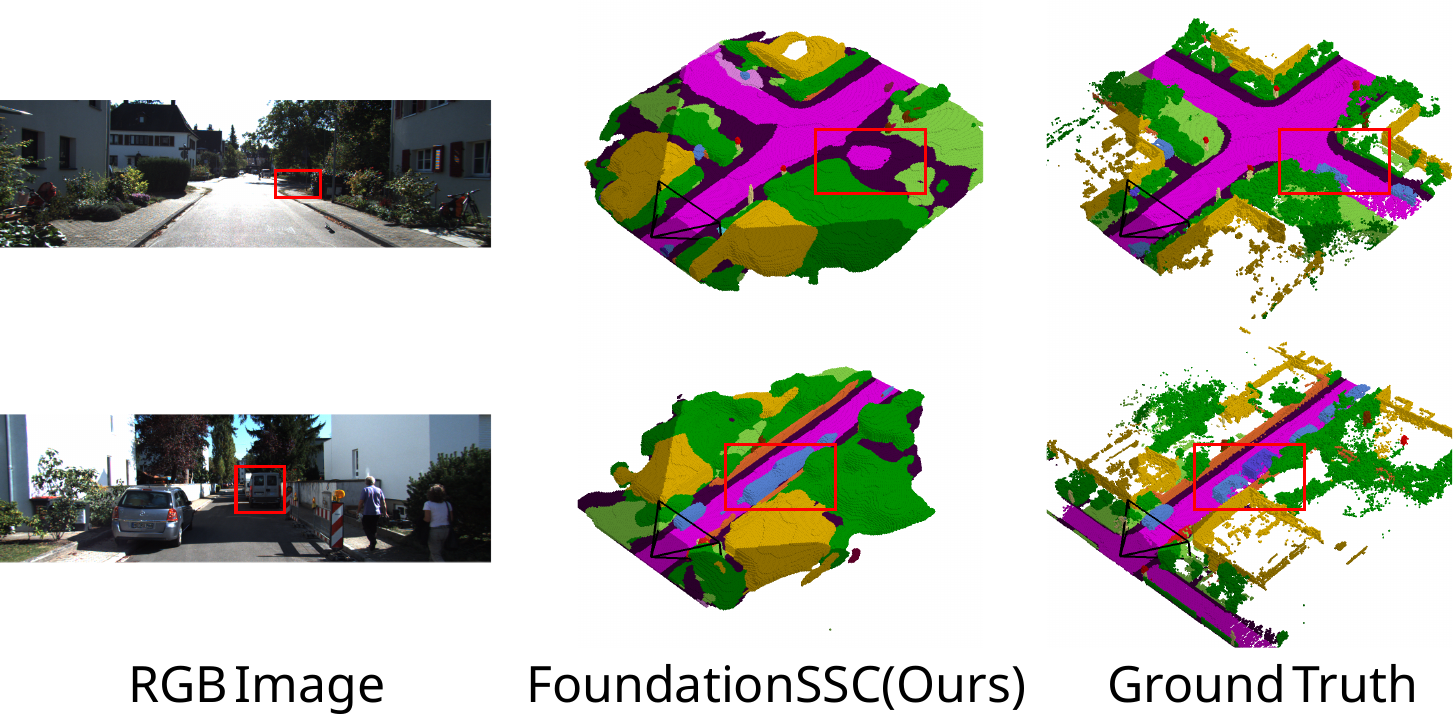}
    \caption{Failure Cases.}
    \label{fig:failure}
\end{figure}

\begin{table*}[ht]
\centering
\newcommand{\clsname}[2]{
    \rotatebox{90}{
        \hspace{-6pt}
        \textcolor{#2}{$\blacksquare$}
        \hspace{-6pt}
        \renewcommand\arraystretch{0.6}
        \begin{tabular}{l}
            #1                                      \\
            \hspace{-4pt} ~\tiny(\semkitfreq{#2}\%) \\
        \end{tabular}
    }}
\renewcommand{\tabcolsep}{2.9pt}
\renewcommand\arraystretch{1.0}
\scalebox{0.8}
{
\begin{tabular}{l|cc|ccccccccccccccccccc}
    \toprule
    Method                               &
    IoU                                  &
    mIoU                                 &
    \clsname{road}{road}                 &
    \clsname{sidewalk}{sidewalk}         &
    \clsname{parking}{parking}           &
    \clsname{other-grnd.}{otherground}   &
    \clsname{building}{building}         &
    \clsname{car}{car}                   &
    \clsname{truck}{truck}               &
    \clsname{bicycle}{bicycle}           &
    \clsname{motorcycle}{motorcycle}     &
    \clsname{other-veh.}{othervehicle}   &
    \clsname{vegetation}{vegetation}     &
    \clsname{trunk}{trunk}               &
    \clsname{terrain}{terrain}           &
    \clsname{person}{person}             &
    \clsname{bicyclist}{bicyclist}       &
    \clsname{motorcyclist}{motorcyclist} &
    \clsname{fence}{fence}               &
    \clsname{pole}{pole}                 &
    \clsname{traf.-sign}{trafficsign}
    \\
    \midrule
    
    MonoScene  & 36.86 & 11.08 & 56.52 & 26.72 & 14.27 & 0.46  & 14.09 & 23.26 & 6.98  & 0.61  & 0.45  & 1.48  & 17.89 & 2.81  & 29.64 & 1.86  & 1.20  & 0.00  & 5.84  & 4.14  & 2.25 \\
    TPVFormer  & 35.61 & 11.36 & 56.50 & 25.87 & 20.60 & 0.85  & 13.88 & 23.81 & 8.08  & 0.36  & 0.05  & 4.35  & 16.92 & 2.26  & 30.38 & 0.51  & 0.89  & 0.00  & 5.94  & 3.14  & 1.52 \\
    OccFormer  & 36.50 & 13.46 &  {58.85} & 26.88 & 19.61 & 0.31  & 14.40 & 25.09 &  {25.53} & 0.81  & 1.19  &  {8.52}  & 19.63 & 3.93  & 32.62 & 2.78  & 2.82  & 0.00  & 5.61  & 4.26  & 2.86 \\
    VoxFormer  & 44.15 & 13.35 & 53.57 & 26.52 & 19.69 & 0.42 & 19.54 & 26.54 & 7.26 & 1.28 & 0.56 & 7.81 & 26.10 & 6.10 & 33.06 & 1.93 & 1.97 & 0.00 & 7.31 & 9.15 & 4.94 \\
    Symphonize & 41.92 &  {14.89} & 56.37 & 27.58 & 15.28 &  \textbf{0.95}  & 21.64 & 28.68 &   {20.44} &  {2.54}  &   {2.82}  &   {13.89} & 25.72 & 6.60  & 30.87 &  {3.52}  & 2.24  & 0.00  & 8.40  & 9.57  & 5.76 \\
    H2GFormer  & 44.69 & 14.29 & 57.00 &  {29.37} &   {21.74} & 0.34 & 20.51 & 28.21 & 6.80 & 0.95 & 0.91 &  {9.32} &   {27.44} & 7.80 &  {36.26} & 1.15 & 0.10 & 0.00 & 7.98 &  {9.88} & 5.81 \\
    CGFormer   &  {45.99} &  {16.87} &   {65.51} &  {32.31} & 20.82 & 0.16  &  {23.52} &   {34.32} & 19.44 &   {4.61} &  {2.71}  & 7.67  &  {26.93} &  {8.83}  &   {39.54} & 2.38  &  {4.08}  & 0.00  &  {9.20} &  {10.67} &   {7.84} \\
    HTCL  &45.51& {17.13}&63.70& {32.48}& {23.27}&0.14&24.13&34.30&20.72& {3.99}&2.80&11.99&26.96&8.79&37.73&2.56&2.30&0.00& {11.22}& {11.49}&6.95\\

    VLScene &44.69	&17.83	&63.10	&31.10	&\textbf{24.40}	&0.20	&24.90	&33.40	&30.70	&1.80	&3.60	&18.30	&26.00	&8.10	&35.30	&4.30	&2.60	&0.00	&12.10	&11.90	&6.30 \\

    Ours &\textbf{47.91} &\textbf{20.36} &\textbf{67.37}	&\textbf{34.12}	&21.70	&0.77	&\textbf{27.96}	&\textbf{37.08}	&\textbf{41.20}	&\textbf{5.94}	&\textbf{7.07}	&\textbf{20.12}	&\textbf{28.61}	&\textbf{10.95}	&\textbf{41.06}	&\textbf{5.02}	&\textbf{4.15}	&\textbf{0.04}	&\textbf{13.01}	&\textbf{12.65}	&\textbf{8.01} \\

    \bottomrule
\end{tabular}
}
\caption{Quantitative results on SemanticKITTI validation set. \textbf{Bold} denotes the best performance.}
\label{tab:sem_kitti_val}
\end{table*}

\begin{table*}[t!]
  \centering
  \small
 \setlength{\tabcolsep}{4.5mm}
  \begin{tabular}{l|cc|ccc}
    \toprule
    {\textbf{Foundation Encoder}} & {\textbf{IoU}} & {\textbf{mIoU}} &\textbf{Params (M)} & \textbf{Trainable Params (M)} &\textbf{Training Memory (MB)}\\
    \midrule
    FoundationStereo-L     &48.12    &19.32   &444.48  &67.38  &22030 \\
    FoundationStereo-S     &47.52    &18.05  &126.04  &61.12  &21400 \\
    \bottomrule
  \end{tabular}
  \caption{Efficiency and performance comparison of FoundationSSC with Large vs. Small FoundationStereo encoders on SemanticKITTI hidden test set.}
  \label{tab:lightweight}
\end{table*}

\subsection{Failure Cases}
Fig.~\ref{fig:failure} presents two challenging scenes that highlight the limitations of FoundationSSC in occluded regions. In the top example, the model fails to infer the presence of an intersection ahead, resulting in incomplete road structure prediction. In the bottom example, one vehicle is partially occluded by another. Due to the lack of instance-level supervision during training, the model struggles to delineate the boundaries of the occluded vehicle, leading to merged voxels between the two cars.

\subsection{Results with Lightweight Foundation Encoder}
To assess the efficiency of our method, we evaluate FoundationSSC using a lightweight backbone by replacing the FoundationStereo-L encoder with FoundationStereo-S. As shown in Table~\ref{tab:lightweight}, this substitution significantly reduces computational cost while maintaining competitive performance on SemanticKITTI hidden test set.

\subsection{Limitations}
FoundationSSC currently operates on single-frame inputs. Although it achieves state-of-the-art performance, it does not leverage temporal information, which could help address cross-frame inconsistencies and enhance overall scene understanding. Incorporating temporal cues will be an important direction for our future work.

\section{Acknowledgments}
This work was supported in part by the National Natural Science Foundation of China (62473276, 62573309), in part by the Natural Science Foundation of Jiangsu Province (BK20241918), and in part by the Research Fund of Horizon Robotics (H230666).


\bibliography{aaai2026}

\end{document}